# Calico: Relocatable On-cloth Wearables with Fast, Reliable, and Precise Locomotion

ANUP SATHYA, University of Maryland, College Park, USA
JIASHENG LI, University of Maryland, College Park, USA
TAUHIDUR RAHMAN, University of California, San Diego, USA
GE GAO, University of Maryland, College Park, USA
HUAISHU PENG, University of Maryland, College Park, USA

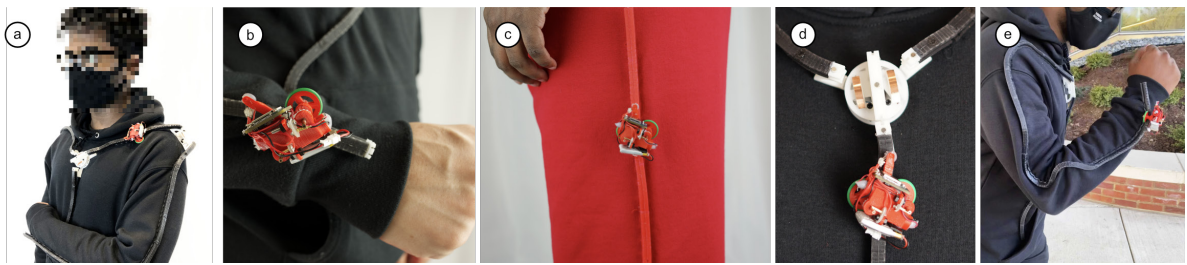

Fig. 1. a) Calico system deployed on a user. b) Calico wearable on the wrist. c) Calico robot moving on the pants. Different colored tracks can be used to blend into clothing. d) Calico moving towards a turntable to switch tracks. e) Running while wearing the Calico system.

We explore Calico, a miniature relocatable wearable system with fast and precise locomotion for on-body interaction, actuation and sensing. Calico consists of a two-wheel robot and an on-cloth track mechanism or "railway," on which the robot travels. The robot is self-contained, small in size, and has additional sensor expansion options. The track system allows the robot to move along the user's body and reach any predetermined location. It also includes rotational switches to enable complex routing options when diverging tracks are presented. We report the design and implementation of Calico with a series of technical evaluations for system performance. We then present a few application scenarios, and user studies to understand the potential of Calico as a dance trainer and also explore the qualitative perception of our scenarios to inform future research in this space.

Authors' addresses: Anup Sathya, University of Maryland, College Park, Department of Computer Science, USA; Jiasheng Li, University of Maryland, College Park, Department of Computer Science, USA; Tauhidur Rahman, University of California, San Diego, Halıcıoğlu Data Science Institute, USA; Ge Gao, University of Maryland, College Park, School of Information Studies, USA; Huaishu Peng, University of Maryland, College Park, Department of Computer Science, USA.

## 1 INTRODUCTION

Wearable devices such as fitness trackers and smartwatches have been widely adopted in our daily lives for both health monitoring and digital interactivity. We are also witnessing the trend of wearable device minimization, where both the size of these devices and their distance from our bodies are reducing [13, 19, 20, 46], suggesting a future where miniature wearable devices may seamlessly integrate with us for interaction, actuation and sensing. However, most wearables to date are resigned to a single location on our body. In reality, the human body offers a multitude of interaction modalities at various places. For example, the collar of one's clothing can be good for voice commands; the forearm offers ample area for spatial feedback and can thus be used for hand input [14]. The areas suitable for sensing also vary. For example, it is best to monitor breathing from the front or back of one's upper body [51], while the wrist can be ideal for sensing activities such as typing and writing. Thus, if a user wants to interact with physiological information or needs haptic or visual feedback from different areas, they are destined to wear multiple wearable devices in various form factors, on numerous areas of the body.

Recent work, such as Rovables [9], proposes an alternative way of thinking about wearable technology, where the locations of these devices can be flexible. For example, Rovables can move on one's clothing to nudge different areas of the torso for discrete feedback; multiple Rovables can park at joints and calculate inverse kinematics, providing on-demand motion tracking. The Rovables project illustrates a wide range of applications where wearables can benefit from having the ability to relocate. However, its locomotion mechanism is not yet ready to realize the full potential of this concept. As Rovables relies on magnetic-rollers that pinch the clothes, it faces multiple challenges that prevent it from travelling over thick clothing and seams that may interfere with the magnetic pairing of the wheels, going around sharp curves such as bent elbows, or grasping on tight-fit clothing which don't provide enough room for the magnetic rollers on the underside of the clothing. The mechanism also places limits on the speed [1] and precision. As a result, the relocatable robots are not mature enough to be deployed in a real-world environment, where the locomotion of the on-cloth robot needs to be fast, precise, and reliable, all without restricting the user's movements.

In this paper, we present Calico, a relocatable wearable prototype with a reliable on-cloth track system to address the aforementioned challenges. With Calico, robots can traverse long distances across different areas of the body, such as from one's forearm to the back, or from the thigh to the chest, regardless of the material deformation or the seams between different pieces of cloth. The locomotion is up to 227 mm/s, about 4.5 times faster than that of Rovables [1], and an accuracy of up to 4 mm. In doing so, Calico can potentially accommodate many common activities by reliably traversing to various locations on the body. For example, Calico can move along the user's arm to provide feedback about their running progress, then move back and forth on the user's back to provide feedback on their posture when sitting down.

The key to Calico is an on-cloth track system — inspired by the railways — on which the wearable moves. In a railway system, rail tracks are used as dedicated pathways to navigate unfriendly terrain; railway turntables are used as central hubs to allow trains to switch tracks. In a similar vein, Calico stitches custom soft tracks directly onto clothing which then function as expressways to overcome the material's natural deformations. Calico also includes rotational switches, like railway turntables, to allow on-cloth track switching and to ensure that all the critical areas of the human body are reachable. As adding tracks to clothing affects its appearance, weight, and performance, throughout the iterative design process, we carefully consider the system's comfort and ensure that the aesthetics of the system can be modified according to the user's needs and preferences.

---

[1]Rovables speed should be no greater than 50 mm/s, which is estimated based on the published supplementary video: https://vimeo.com/206134759.

In the rest of this paper, we first detail the design consideration and the system implementation of Calico, with a series of technical evaluations for its performance. Along with the system design, we present a suite of application scenarios to illustrate the potential use cases in which Calico can be utilized, including remote end-user auscultation, fitness coaching, a dance trainer, and a personal assistant that can physicalize data. We follow this with a lab study (N = 14) to evaluate the potential usefulness of Calico while learning a dance move. We also present a scenario-based survey study (N = 50) to understand the perception of our system in distinct scenarios to aid future research in this space.

In summary, we see Calico's main contributions as the following:

(1) A track-based relocatable wearable design that allows on-cloth wearables to move in a quick, precise, and reliable manner. For the first time, a wearable can traverse clothing with no restrictions to the user's movement.
(2) A suite of applications and scenarios that demonstrate the promise of on-cloth wearables using a track system.
(3) A lab study to evaluate Calico in a dance learning scenario, and a survey study to understand the perceived usefulness, joyfulness and concerns about the on-cloth wearable concept.

## 2 RELATED WORK

Our work builds upon the concepts of on-cloth wearables, on-body interaction and sensing, and locomotion mechanisms for relocatable wearables.

### 2.1 On-cloth Wearables

Clothes, being essentials that come into contact with many parts of the human body, have naturally been studied substantively in HCI and used as platforms for novel interactions [23, 36–38, 41, 44, 48]. For example, various touch input methods have been proposed using sleeves [37, 38, 44], pockets [43], button snaps [10], headphone cords [36], and zippers [26]. As most on-cloth wearables have a minimal device size and are mounted to the garments rather than attaching them to the body itself, they often provide a slightly less intrusive method for interaction and sensing. Still, on-cloth wearables require sensors to be fixed to certain locations on the clothing, thereby limiting the interaction and sensing space.

### 2.2 On-body Interaction and Sensing

One important application for wearable devices is on-body interaction and sensing. In the field of HCI, wearable devices have been used to offer haptic feedback at various locations. For example, several researchers propose the use of haptic feedback for spatial information using skin-drag [17] or pin-based feedback mechanisms [16] at the wrist. Full-body haptics has also been explored using vibration motors [31] or pneumatic pockets [8], where a number of actuators need to be instrumented as part of the wearable device.

Researchers have also explored bio-signal sensing and on-body interactions at multiple locations on the human body [51]. For example, work has been done to detect touch coordinates or distinguish between ways of typing, by placing bio-acoustic [14, 29] or electromagnetic sensors [30, 52, 53] near the wrist. Considerable efforts have also been made towards recognizing human activities at different locations of the body [4] — daily activities such as sitting, standing, walking, ascending & descending stairs (shirt pocket, belt) [34], eating & drinking (neck, wrist and upper arm) [1, 2, 40, 49]; sports like basketball (wrist) [15, 35], cricket (all 4 limbs) [24] and weightlifting (waist) [5].

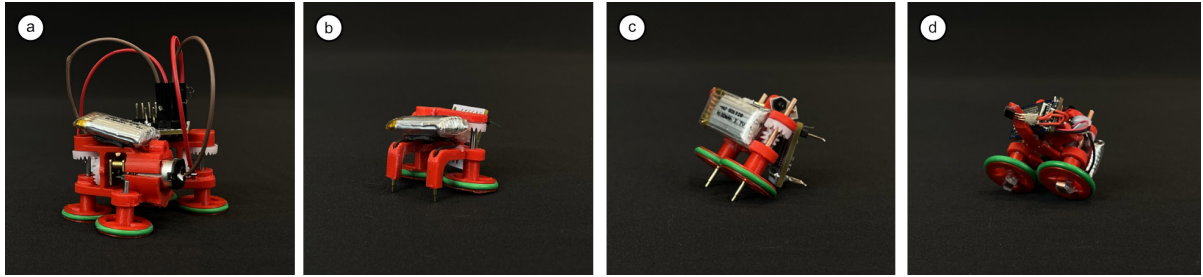

Fig. 2. The different design iterations of Calico. a) The proof-of-concept with 4 wheels and 2 Pololu Micro Metal Gearmotors. b) Switching to a smaller motor with a different gear assembly. The overhangs would easily get caught on wrinkles in the clothing. c) Switching to a copper shaft to avoid the overhangs but still create electrical contact with the turntable. d) Switching to dual motor drive and magnets with copper plates to increase the speed and the climbing capacity.

As we move towards a future where context awareness and whole-body feedback becomes more prevalent, apart from enabling multiple applications, Calico also creates new avenues for continuous feedback along the body instead of the discrete feedback which is currently standard.

### 2.3 Locomotion Mechanisms for Relocatable Wearables

In recent years, researchers have proposed new relocatable wearables, where sensors and actuators are not fixed to one place, but can be relocated. For example, Rubbot [6], Clothbot [33], CLASH [3], and Rovables [9] employ different types of cloth-pinching mechanisms that allow wearables to move freely on cloth substrates without modifying the material. Such free climbing mechanisms, however, come with trade-offs. To avoid falling from the human body, these devices need to firmly grasp and deform the fabric, making locomotion slow and inaccurate. Additionally, as the clothing needs to be deformed, it can only be deployed on loose and thin fabric. In other cases [9], a magnetic wheel is placed under the clothing to allow the robot to latch onto the fabric. This underside magnetic wheel constantly touches the skin of the user and also renders the robot incapable of moving across seams, thick fabrics or tight clothing. These devices also rely on dead-reckoning algorithms to localize themselves on clothing, which in turn limits the localization performance.

To avoid these issues, we decided to alter the clothing in a careful and non-intrusive manner to reap the benefits of increased locomotion and localization performance. To the best of our knowledge, the only work that is similar in concept is [42], in which researchers attach a single belt to the human's arm as a robot rail. However, their implementation is preliminary, as the rat-sized robot only travels in a single dimension along the arm. In our design, we aim to drastically reduce the size of the robot, make the movement much faster, increase the number of areas to which the robot can move, and also improve the social acceptance of the system by providing mechanisms for aesthetic flexibility.

## 3 CALICO SYSTEM

We now detail the design and implementation of our system. We start with our design considerations.

### 3.1 Design Considerations

Several key considerations and constraints dictated the design of the Calico system.
- **High speed**: Wearables may be used all across the human body. Thus, it is critical that our device can move quickly in order to be responsive and to accommodate different scenarios on time.

- **Robust movement**: The human body is rarely stationary. Thus, our device needs to remain functional and stay on the clothing when the user performs different activities such as walking, jogging, jumping or dancing.
- **High reachability**: Since multiple body areas can be useful for sensing and feedback, our system should be able to reach a large number of critical areas such as the chest and back, shoulders, arms, wrists, and legs.
- **High precision**: Some applications, such as chest auscultation or breathing detection, require the device to move precisely to a certain location for accurate data collection. Thus, it is essential that the wearable knows its current location in relation to the human body and can reach the destination in a fine-grained manner.
- **Low weight and small form factor**: Since wearables will be worn daily, it is important to maintain a small footprint with a low weight to minimize the burden placed on your body.
- **High payload capacity**: Different applications may require the use of various sensors and actuators. It is ideal to have a payload capacity high enough to carry sensors and components to handle such applications.
- **Aesthetic flexibility and social acceptance**: As an on-cloth wearable device, we should also consider its social acceptance. We aim to make the system unobtrusive, comfortable to wear, and also create room to modify its aesthetics.

Along with the above design considerations we also carefully weighed the recommendations provided by previous work based on user perception and surveys [18, 21, 51].

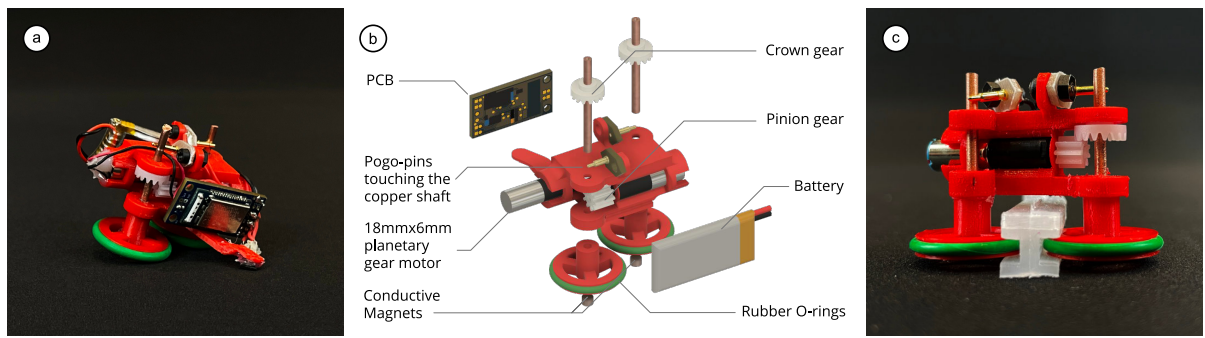

Fig. 3. a) The Calico bot. b) An exploded view of all the components that make up the Calico bot. c) The latching mechanism used by the Calico bot to move along the track.

## 3.2 The Robot

Figure 2 presents the early prototypes of the robot with an iterative design process. Our design is centered around the idea of a "railway system" on cloth — Calico's core climbing mechanism. To ensure reliable climbing while minimizing the impact to the wearer, we also designed the robot so that the wheels do not have direct contact with the human body or the clothing, but latch onto the track system directly.

Our initial prototype had four wheels to ensure a stable grasp of the track (Figure 2a). The robot had a ball bearing at its center, which allows the front and back sections to pivot in one dimension around the central axis. Hence, the robot can move along tracks with curves in parallel to its chassis. However, the four-wheel configuration does not have the flexibility to freely rotate in both dimensions around its central axis, preventing the robot from going around curves that are normal to its base, such as those generated around the elbow or while moving from the arm to the shoulder. As it has four wheels, the size of the robot is also relatively large.

Figure 3 shows the final design of our Calico robot. It uses the same wheel-track pinching mechanism as the initial prototype but employs two independently driven wheels. Each wheel has a diameter of 18 mm and is covered with a rubber O-ring to increase the grip between the wheels and the track. Two 242 rpm, 6 mm planetary gear motors, and an additional 2:3 ratio crown gear set are used in concert to drive the two wheels. We use two motors to ensure that the robot has enough torque to overcome gravity while also providing a high payload capacity. The crown gear allows us to install motors parallel to the robot's wheels, which reduces the robot's overall height and lowers the center of gravity of the robot towards the body, making it less likely to sag from the body. Each wheel has a copper shaft to create an electrical connection between the motor driver and the bottom of the wheels. At the bottom of each wheel, we included a neodymium magnet covered with a copper leaf (Figure 2d). These act as contact points to drive the turntables, which we will detail in Section 3.4.

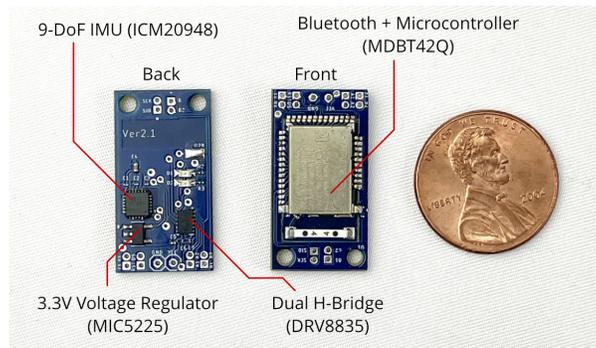

Fig. 4. Custom PCB designed to maintain a small device footprint.

As Calico is designed with minimization in mind, we created a custom 2-layer PCB to control and communicate with the robot (Figure 4). The PCB is 13 mm × 24 mm, with an MDBT42Q on one side (built on the NRF52832 by Nordic Semiconductors) as the main micro-controller. On the other side, a DRV8835 dual H-Bridge and an ICM20948 9DOF IMU are used to drive the motors and for on-board sensing, respectively. The MDBT42Q has native support for Bluetooth Low Energy (BLE), which is Calico's primary communication mechanism. We also include two A3144 hall-effect sensors at the bottom of the robot used for on-track localization and controlling the turntable, which will be discussed further in Sections 3.3 and 3.4. The robot is powered by a 3.7 V 100 mA h Li-Po battery with a size of 30 mm × 4 mm × 13 mm. Once assembled, the Calico robot has a size of 42 mm × 32 mm × 35 mm, and a weight of 18 g. It can carry a 20 g payload — more than its body weight — and achieve a speed up to 227 mm/s. We will detail the experiments conducted to evaluate the robot's performance in Section 5.

### 3.3 The Track

Figure 5 is our custom track design. From the wearer's perspective, the track needs to be lightweight and not present a burden when attached to clothing. It also has to be flexible to ensure that the movement of the body is not restricted. From the system's perspective, the track needs to be deformable so that the robot's wheels have a sufficient contact area to create enough friction to ensure stability. It also requires a rigid track cover to prevent the robot from derailing. Furthermore, the track has to be easy to stitch and be highly durable to withstand standard clothing care procedures.

Figure 5a is the section view of the track. A 3 mm height, 3 mm thick flexible center wall is sandwiched between two flat panels, one with 4 mm × 8.5 mm at the top and one with 2 mm × 8.5 mm at the bottom. When climbing, the two wheels of the robot will pinch the center flexible wall. The top panel is designed to prevent the robot from running over the track while still remaining flexible along all axes. When the track is bent, the top panel ideally guides the wheel around the track and prevents it from squeezing through and derailing. The flat bottom panel offers ample space for stitching.

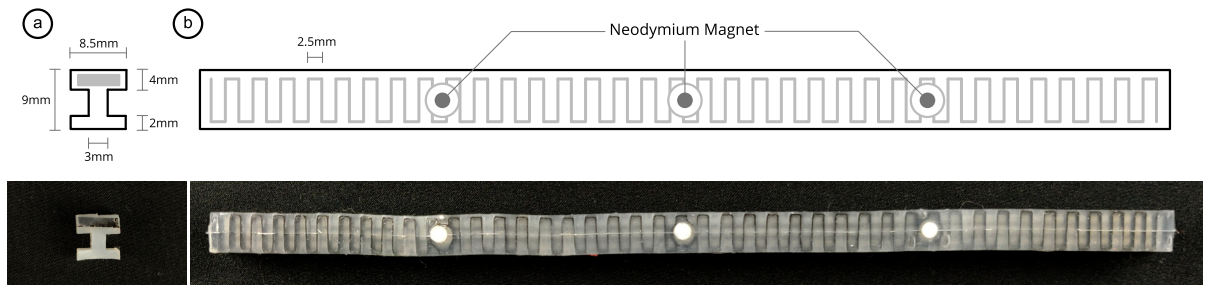

Fig. 5. Meta-material track design. a) Cross-section of the track. b) Top view of the track with embedded magnets and a live-hinge pattern of the inlay.

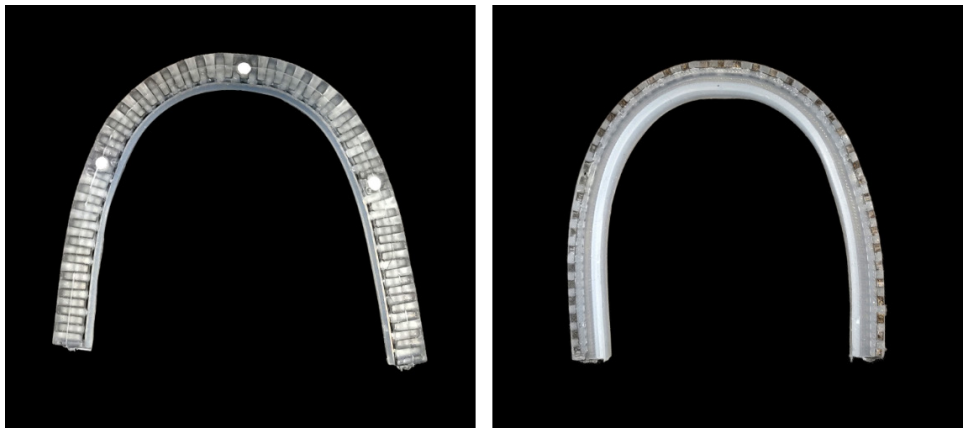

Fig. 6. Due to the live-hinge design, the track can be flexed in all directions while still not allowing the robot to derail from the track.

The major part of the track is molded with platinum catalyzed silicone (Smooth-On's EcoFlex™ 00-50). While experimenting with the molded track, we noticed that the top panel by itself is not sufficient to prevent derail. In fact, the robot would always deform the track and pinch through the top panel, regardless of its thickness. To solve this, we propose a meta-material track design, where the top panel is inserted with a solid inlay, made with a live-hinge pattern (2.5 mm pitch) (Figure 5b). The live-hinge inlay maintains the track's flexibility while offering a rigid cross-section that the wheels cannot penetrate. This inlay is 3D printed with thermoplastic and can almost entirely blend into the track when printed with the proper color (Figure 31).

One benefit of having custom tracks is that we can conveniently include landmarks for precise localization. Here, we embed additional cylindrical neodymium magnets (3 mm × 2 mm) into the track's top panel with a fixed distance of 50 mm between them. This distance can be increased or decreased if needed. When the robot moves over one such magnetic landmark, the twin Hall effect sensors installed at the bottom of the robot can pick up the signal.

## 3.4 The Turntable

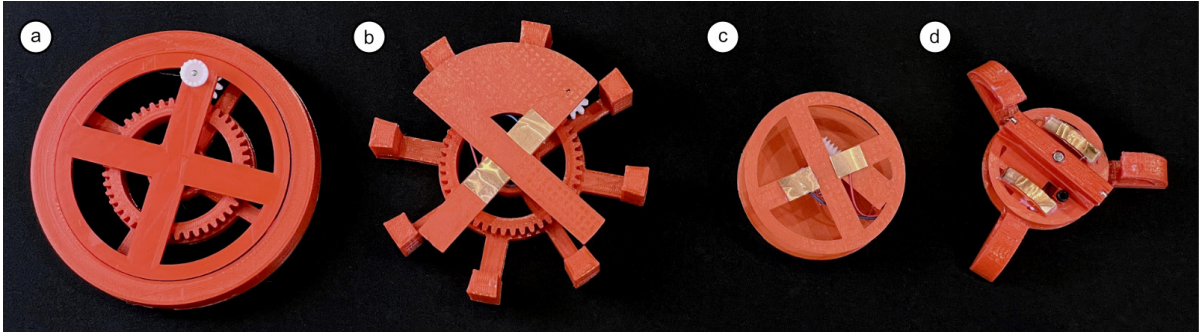

Fig. 7. From left to right, all the major design iterations of the turntable. a) An extra motor on the robot would latch to the crown gear on the turntable to rotate it. b) Moving the motor from the robot to the turntable to avoid meshing gears as creating electrical contact is easier. c) Reducing the size of the turntable by rotating around an external spur gear instead of an internal spur gear. d) Using a gear train to reduce the size further and adding fixed ramps to allow the robot to reliably move onto the turntable.

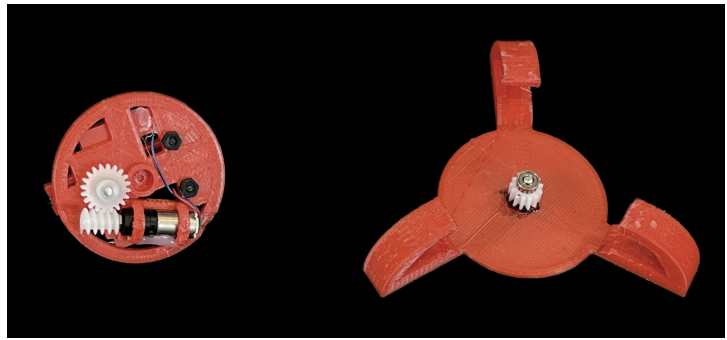

Fig. 8. The turntable mechanism. When the robot makes contact with the copper pads on top and sends a signal through the motor driver, the rotor rotates around the base, which also rotates the robot.

The turntable is the final component of the Calico system. It allows the robot to switch between tracks when diverging routes are presented. Similar to the robot design, we went through multiple iterations (Figure 7) to minimize its size, increase its effectiveness and reduce the number of electronic components on the robot.

Figure 7d and Figure 8 detail the final turntable design. The turntable is composed of two main components — the base with a pinion gear at its center and a rotor sitting on its top. The rotor's bottom side holds an off-centered spur gear, which engages the pinion gear of the base, and a worm gear that is driven by a 1000 rpm 6 mm planetary gear motor. The motor's two electrodes extend to the top side of the rotor, with two copper contact leaves and neodymium magnets underneath. When the robot moves onto the small turntable, the bottom of its wheels, which are equipped with a small neodymium magnet and covered by a copper plate (Figure 2d), will magnetically attach to the turntable's contact leaves. The robot's on-board battery and the motor driver will then activate the rotor's motor, which rotates the turntable along with the robot itself. The Hall effect sensor at the bottom is used to check the orientation of the robot to ensure that it exits in the right direction.

Our final turntable design has a diameter of 40 mm, excluding the ramp. The base of the turntable can be directly stitched on to the clothing while the rotor top that holds the gear motor is removable. This design reduces the effort for maintenance, as the entire cloth becomes washable once we take the top rotors off. The turntable's design allows the inclusion of up to 4 entry and exit points, but for our implementation we utilize 3. The Calico robot can switch tracks with a high speed when on the turntable. A full 360° rotation of the turntable takes only 0.4 s.

### 3.5 Generalized Track Layout

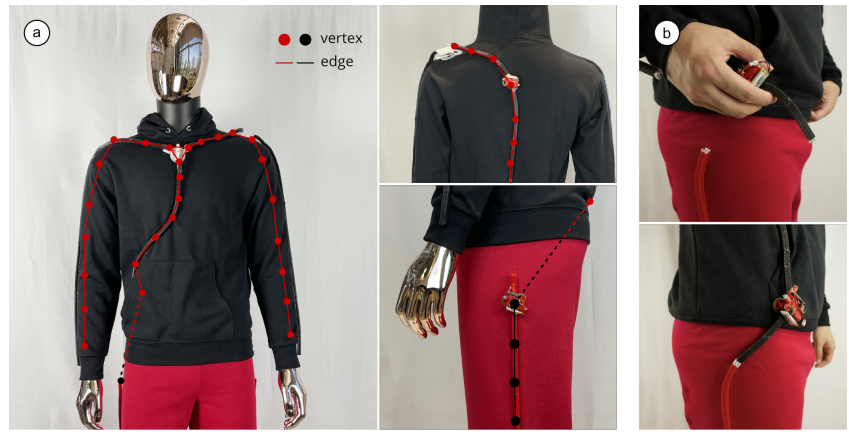

Fig. 9. a) Using a track system on the clothing allows us to project the 3D terrain onto a 2D graph structure to simplify localization. Dotted line indicates temporary magnetic connection. b) The user magnetically connects the tracks to allow the robot to move between upper body and lower body.

We considered two approaches for the deployment of the Calico track system on clothing:

(1) *Use a generalized track layout to support a multitude of applications.* The track can be installed on a commonly used piece of clothing that is socially acceptable in different settings, for example, a dark-colored hoodie that can be worn for a wide range of activities both indoor and outdoor.
(2) *Create application-specific track layouts on context-specific clothing to reduce layout complexity.* For example, an evening dress can contain only enough track to pull a scarf around the user's neck when the temperature decreases. This allows the robot to be faster as it doesn't have to traverse through a complex track layout to perform a simple function in situations where other functions might not be needed.

In this paper, we focus on the former approach to design a generalized track layout on which Calico robot is deployed. This allows us to explore a wide range of applications and user interactions with an unified track design. The latter approach can be tailored to application-specific demands or users with different needs and preferences. However, designing such specialized layouts would greatly benefit from a co-design approach involving relevant stakeholders, which is outside this paper's scope.

Our generalized track layout design tries to maximize the key body areas that can be reached by the wearable. The track layout follows suggestions from [25, 51] to cover several key areas of interest for on-body interaction and sensing. As shown in Figure 9, our track deployment includes 48 points of interest spread across both arms, both legs, the chest, the back, and the stomach area. At two locations — left shoulder and upper chest — where sections of track were coinciding with each other, we install turntables that allow the Calico robot to switch between tracks. As the upper and lower body tracks are on two different pieces of clothing, we further include a magnetic connector at the end of the upper body tracks (Figure 9b). If needed, the user can snap this magnetic connector onto the corresponding entry point on the pants to allow Calico to traverse between the upper and lower body.

### 3.6 Interaction, Actuation, and Feedback

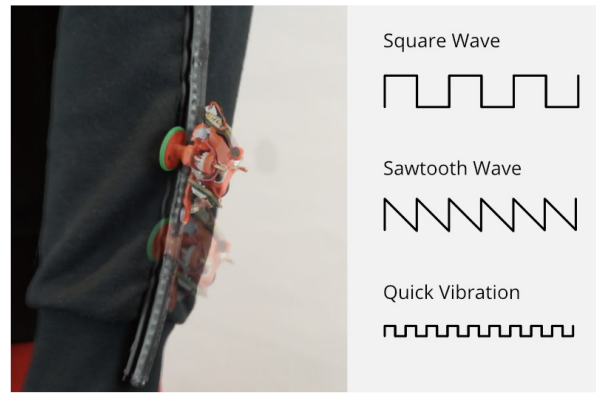

Fig. 10. The different forms of vibration patterns that Calico uses to communicate with the user.

With the deployed track system, Calico robot is capable of understanding its position on the user's body by detecting physical landmarks on the track (Section 4.1). The on-board IMU can sense haptic input directly from the user. For example, the robot can sense if a user is touching or tapping its back. The prototype can be further expanded with additional components when further functionality is needed. Section 6 details one such example where a miniature microphone is attached to capture the user's breath. The PCB can also accommodate a capacitive touch surface.

Calico is also capable of giving different forms of feedback to the user. For example, the on-board LED can communicate basic robot status with simple flashing patterns. It can also provide more expressive haptic feedback by moving along the track in different patterns — square wave, sawtooth wave and a quick vibration (Figure 10). Compared to haptic feedback from a stationary wearable such as a smart watch, the movement patterns of Calico can be location-specific. It therefore provides us context-specific haptic feedback similar to what can be called "Body-Centric Interaction" [7, 47].

### 3.7 Beyond On-cloth Interaction

Besides full-body interactions, the use of a track system makes it possible to relocate, or offload the Calico robot beyond the user's body. When the user doesn't need the Calico robot at that particular moment, they can connect the end of the track from the sleeve to a wall-mounted key hanger (Figure 11). Calico robot can then move from the user to the wall, to stay among the rest of the user's belongings such as keys. Additionally, a Calico robot can be lent from one user to another. As shown in Figure 11, two users can snap the ends of the track on their wrists together using the pre-attached magnetic connectors. The Calico robot can then move across the gap and transfer from the original host to another user. The new user can then control the Calico robot using their smartphone. Similar concepts have been explored as "Parasitic Mobility" [27] and also for smartphone sharing [22, 32, 45] ; we envision that robot sharing can be relevant when the Calico platform matures.

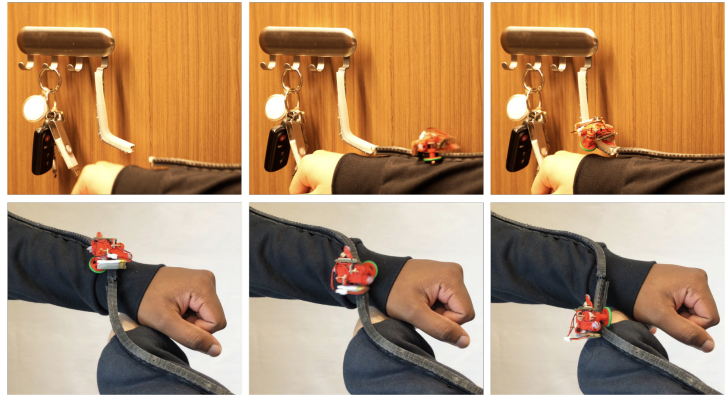

Fig. 11. *Top* — Using magnetic connectors at the end of the track, Calico can move from the user to a key hanger at the end of the day. *Bottom* — This functionality can also be extended to other users.

## 4 IMPLEMENTATION

The track requires a custom fabrication and installation procedure that we detail in this section. We also detail the software and the protocols used to communicate with the robot, move it to a particular location and localize it on the body.

### 4.1 Fabricating the Track

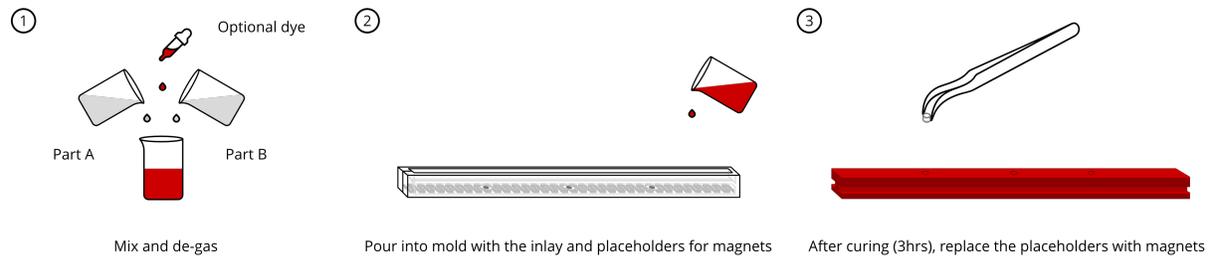

Fig. 12. 1) The 2 part platinum-catalyzed silicone is mixed and degassed. 2) The silicone mixture is poured into a reusable 3D printed mould containing the inlay and placeholders for magnets. 3) After allowing to cure, the track is removed from the mould and the placeholders are replaced with neodymium magnets.

The track is fabricated using a 3-step procedure (Figure 12). *First*, the two part EcoFlex is mixed and de-gassed. An optional dye can be added at this stage to change the color of the track. *Second*, the silicone mix is poured into the 3D printed mold with a live-hinge inlay and placeholders for magnets. *Third*, the track is cured for 3 hours. After this curing process, the magnet placeholders are replaced with the cylindrical magnets and fixed using a silicone epoxy (Smooth-On SilPoxy™). One piece of such track is 200 mm long. We can connect multiple modular tracks with silicone epoxy for the desired length. The track can also be trimmed into smaller sections for reduced length.

## 4.2 Installing the Track

Stitching the track directly onto clothing using polyester thread and a whip-stitch pattern serves as a robust long-term deployment method. We also tested the possibility of using industrial velcro. The installation procedure is cumbersome and unreliable as the silicone track does not adhere to velcro using standard adhesive and thus has to be stitched to one side of the velcro, which is stiff and hard to penetrate. In our implementation, we hand-stitched the track onto clothing. A 200 mm section of track can be stitched by a novice in 5 minutes. The whip-stitch pattern allows the track to maintain flexibility in all the required directions while also being robust and maintaining a nice edge without unraveling the fabric. Some modern sewing machines can handle whip stitch patterns but as we modified already fabricated consumer clothing, this method was unsuitable. Machine stitching can be easier to implement before panels of cloth are turned into clothing items.

*4.2.1 Durability.* As the track is fabricated with silicone, its heat resistance is high enough for at-home cleaning and drying procedures. The PETG implant is also protected by the silicone moulded around it, thereby preventing any damage. This allows the track to be fully washed and dried using a consumer washer and dryer.

## 4.3 System Control and Localization

*4.3.1 Control.* The MDBT42Q is flashed with the Espruino[2] bootloader, a JavaScript interpreter for low-power devices. This low energy consumption is achieved by being entirely event-driven. The Calico robot connects to a BLE-enabled device such as a smartphone or a laptop with all communication and control running via a web browser or a web based IDE. With low energy consumption, Espruino's MTU (Maximum Transmission Unit) is 20 bytes. These constraints could have implications for applications that require high bandwidth.

*4.3.2 Localization.* The track is embedded with small cylindrical neodymium magnets as explained in Section 4.1. The Hall effect sensors on the robot allow us to detect these magnets, with which the robot's location can be projected onto a 2D graph structure (Figure 9). As each unique location triggers the Hall effect sensor in the same way, the initial position of the robot is entered by the user. After the initial position is noted, further localization is done by dead-reckoning [12]. As shown in Figure 9, the magnets form the vertices (V) and the pieces of track connecting the vertices forms the edges (E). This graph structure is built and stored on the controller (smartphone, laptop). As a result, the robot itself does not directly interact with the graph structure, but relies on the controller to navigate the robot. Adding more magnets to a section can increase the localization resolution but this poses a few challenges described in Section 8.3.

The 2D graph structure is traversed by the shortest path using Djikstra's algorithm [11] — $\Theta((|V|+|E|)log|V|)$. As the robot can move in two directions, both up and down, Djikstra's algorithm alone is insufficient to determine the direction in which the robot should move. We solve this by always storing the previous vertex visited by the robot. When a new path is received, a search algorithm is run to determine if the previous vertex is present in the new path. If the previous vertex is present in the new path, the direction of movement is reversed.

## 5 PERFORMANCE BENCHMARKS

We now report a list of evaluation experiments aiming to understand the performance of the Calico prototype. For all the experiments, we designed and built test beds to evaluate a chosen parameter.

## 5.1 Speed

The speed of Calico can be influenced by multiple parameters — the friction between the wheel and the track, slippage, and gravity. Intrinsically, the robot is slower while going up a positive incline and faster going down a negative incline. Adding a payload should also affect the performance.

---

[2]Espruino Bootloader - https://www.espruino.com

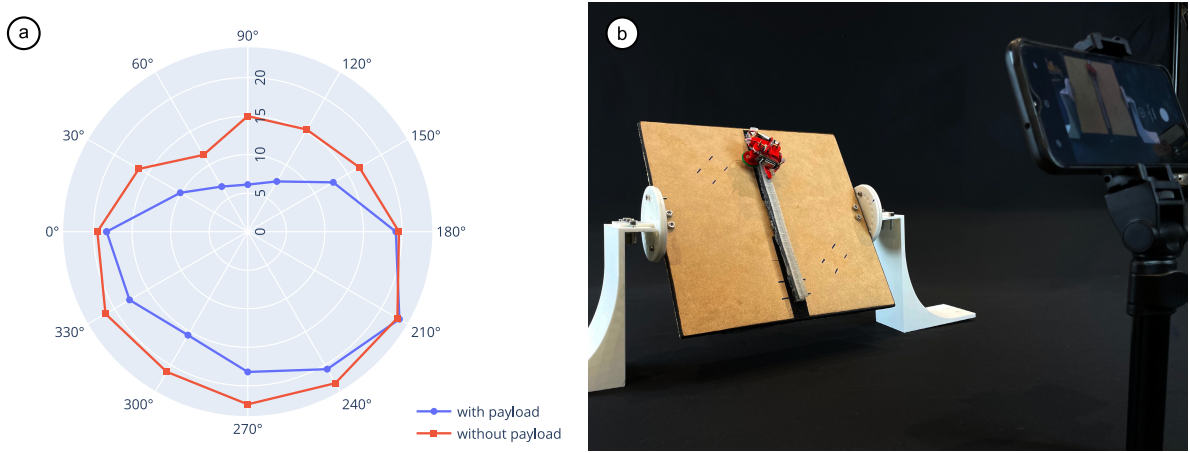

Fig. 13. a) Speed in $cms^{-1}$ with and without a payload (20 g) plotted against approach angle. b) Test setup used to test the speed with and without a payload.

To test the effect of the meta-material track on the climbing capability at different angles, we built an angular testbed as is shown in Figure 13b. This testbed allows us to adjust the approach angle in 15° increments across the full 360° range. A piece of track is attached to the testbed with markers for the start and finish line 140 mm apart. The speed is calculated by filming the bot at 240 fps and counting the number of frames it takes to go from the start line to the finish line. The average readings for 3 trials per angle in 30° increments are recorded in Figure 13a.

The track and the weight distribution of the device presents multiple interesting results. For instance, the robot is fastest while going down a negative 120° incline rather than a negative 90° incline. The center gravity of the robot is located away from the track and the body. As a result, while going down a straight drop, the robot flexes the track around the wheel into a shape that increases the contact area and adds additional friction. This flex is absent while going down a negative 120° incline. Similar behavior is seen while going up a positive incline.

As we can see, with the current configuration of the robot, the device moves at 115 mm/s at worst and 227 mm/s at best. For instance, if we imagine an application where the bot moves from the wrist to the shoulder with an on-demand microphone required for better audio quality, such relocation can be achieved in about 4 s, assuming that the robot is moving straight up. This speed affords the implementation of real-time interactive applications involving the robot.

## 5.2 Maximum Payload and Speed under Payload

We also tested each approach angle with increasing payloads using a set of test weights. We found that the payload that the device could carry was an extra 20 g across the full 360° approach angle range. Hence, we re-tabulated the speed to examine the effects of the 20 g payload (Figure 13a). As expected, the payload only mildly affects the performance of the robot while going down a negative approach angle as the robot is helped by gravity. At worst, the speed is reduced by 60%.

With a 20 g payload (test weight attached using industrial Velcro), the flexing behavior explained in the previous section is shifted by a few degrees as the payload moves the center of gravity of the robot further away from the track and sags, thereby flexing the track a little more. For example, the robot is slowest while going up a 60° incline nonloaded, but with a 20 g payload, a 90° incline is the new lower bound of the speed. During the test, we also noticed that inconsistencies with the fabricated track might create hot spots at which the robot sometimes struggles. A few tracks fabricated in an early procedure had air bubbles where the robot would get stuck without flexing the track and going around it. The extra stress added by a payload highlighted these inconsistencies which forced us to reconsider the fabrication procedure. We tuned the shape of the inlay to allow the silicone to settle in more easily without creating any air bubbles.

### 5.3 Slip Due to Acceleration

To investigate whether daily activities, such as jumping, sprinting, or arm-swinging, may create potential slip, we built a test bed that can create controllable angular acceleration (Figure 14). A piece of cloth with an attached track was fixed to a rotating rod belt driven by a DC motor. We marked a start position and fixed the robot to that location and then tested the slip at various speeds with g-force automatically logged with the robot's on-board IMU. For each motor speed, we allowed the robot to spin for 5 seconds before noting the slip. Note that for this study, we altered the on-board accelerometer to read at full-scale range (±16 g). We calculated the magnitude of the g-force using the Euclidian Norm.

Figure 14a shows the results of the experiment. The Calico robot undergoes minimal slip below 15 g and starts to slip off completely when the g-force is over 16 g. Above 20 g (marked in red), it tries to go through the 3D printed inlay instead of slipping on the groove. According to the WISDM dataset[3], daily activities (including exercise) generate a maximum of 9 g on the wrist. Calico only starts to slip at around 16 g.

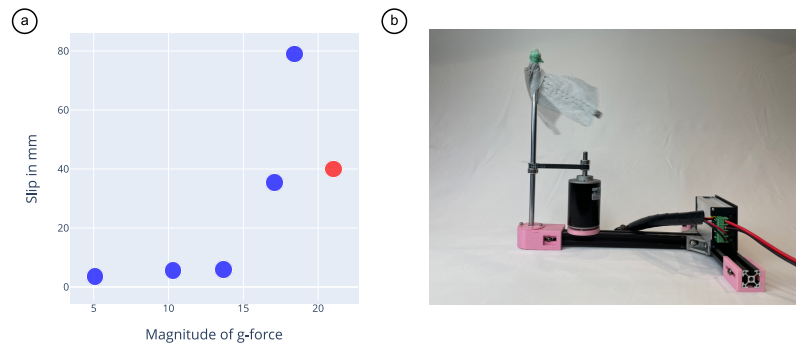

Fig. 14. Testing the slip of the device when subjected to different g-forces. a) Scatter plot showing the slip vs the g-force encountered. At 21 g (marked in red), the wheels broke through the inlay instead of slipping along the groove in the track. b) Test setup used to run the experiment.

### 5.4 Minimum Curvature

As the robot is deployed on the human body, the other main traversal challenge is for the robot to go around curvatures effectively. To ensure the robot can move on all kinds of body terrains, the minimum curvature that the robot can go around has to be satisfactory. This was tested using a custom fixture with a series of printed semicircles installed on top (Figure 15). Scenarios such as these prompted the inclusion of the 3D printed inlay. We started the experiment with a 69 mm diameter rigid semicircle (including the thickness of the track) and continued to reduce it until the robot could no longer go around it. The robot's ability to go around such curves is contingent upon how the wheels flex the track. When the curve becomes too small, the contact area increases to a point where the friction is too high to overcome and flex the track in a convenient direction to allow the robot to move along it.

---

[3]WISDM Dataset: https://archive.ics.uci.edu/ml/datasets/WISDM+Smartphone+and+Smartwatch+Activity+and+Biometrics+Dataset+

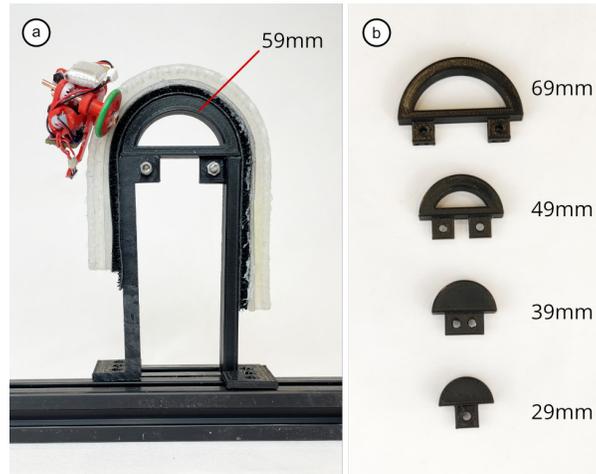

Fig. 15. a) Test setup to determine the minimum rigid curve that can be traversed by Calico. b) Different curvatures used.

We found that the minimum diameter for a convex rigid semicircle was 39 mm. Although concave curves are much less common (armpit, kneepit, inner elbow), we tested the minimum diameter that can be traversed and found it to be 49 mm. This is slightly higher than the convex curve as the body of the robot will also have to accommodated within the concave curve. As the track is attached to clothing, the robot can further deform the track to go around small curves. Nevertheless, care must be taken to ensure that extremely sharp curves are avoided while designing the track system.

5.5 Accuracy

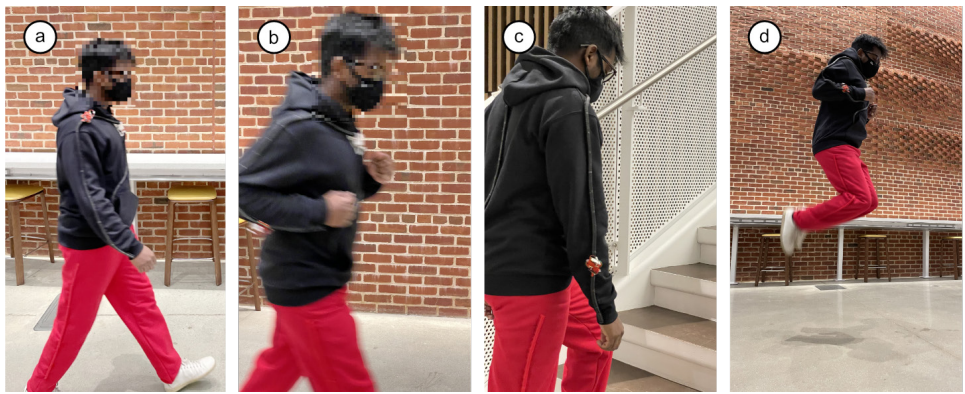

Fig. 16. Testing the navigation of the Calico system under standard user activity.

Although the previous experiments establish the performance of Calico in controlled conditions, the accuracy of the navigation in real-world scenarios can be further examined. Calico's ability to reach a specific anatomical location on the body depends on two factors: 1) how well the Hall effect sensor can detect the magnets embedded into the track and 2) the error between the magnet on the track and the desired anatomical location on the body.

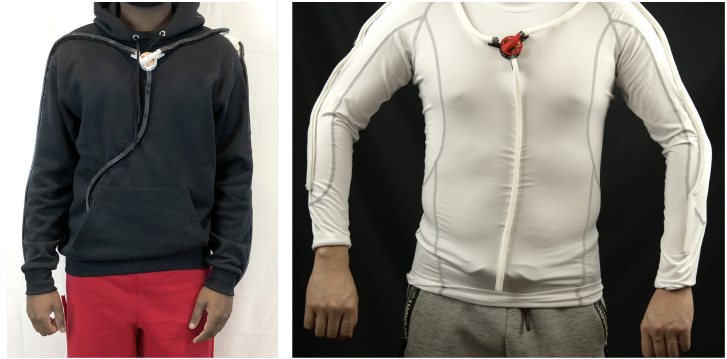

Fig. 17. *Left* - Calico deployed on loose clothing. As the clothing sways away from the body, the track system and the robot moves away from the skin resulting in lower anatomical on-body localization accuracy. *Right* - The tracks and the robot stay close to the body despite the user's movement which translates to higher anatomical on-body localization accuracy.

As the Hall effect sensor provides a closed control loop, the only room for error is if the sensor misses a magnet while moving along the track. Additionally, this variable is limited by the size of the magnet and the clearance in between. For a 3 mm × 2 mm magnet, the minimum resolution is 4 mm (1 mm clearance between magnets). In order to evaluate this, we performed a test to understand whether Calico mis-navigates under standard expected user activities. A representative user from the development team performed three activities — walking, climbing stairs, and jogging (Figure 16) — for 5 minutes each. The user also jumped a few times towards the end of the test. A Calico robot was deployed on the user's clothing and was programmed to travel back and forth between the left wrist and the lower back continuously while he performed the activities. This loop took 16 seconds and included 16 magnets. Throughout the experiment, Calico didn't miss any magnets, thereby accurately moving and staying on loop. This seems to provide a satisfactory localization accuracy with robust locomotion for constant usage even while the user was moving freely. It can also be informally noted that during the entire development cycle, a magnet was never missed by the robot.

The second factor — the error between the magnet on the track and the desired anatomical location on the body — is a function of the tightness of the clothing (Figure 17). Tighter clothing reduces the displacement of the magnet when the user's body moves. When tight compression clothing is deployed, there is no observable displacement between the magnet and the user's body even when the user moves. For applications that require highly accurate anatomical localization, tighter clothing can be preferred and for applications where the accuracy is less important, looser clothing should suffice.

### 5.6 Power Consumption

Although our prototype is not explicitly designed for low power consumption, we aim to extend the operation time if possible. The microcontroller draws 3 μA at its default idle state, and 12 mA when searching for devices to connect. Similarly, the IMU drains 8 μA when idles and 3.11 mA at worst. The two on-board motors consume the most energy — with ~80 mA each during normal climbing. Other components, such as the motor driver and Hall effect sensors, drain less than 1 mA.

In an extreme case, where the robot continuously travels along the user's body with the IMU reporting readings every 100 ms, Calico can have a 30-minute battery life with the 100 mA h battery we are currently using. When it's idle, Calico can last more than 8 hours. To further expand the operation time, we can include a wireless battery charging module, ensuring full-day usage.

## 6 CALICO SCENARIOS

In this section, we use four scenarios to emphasize one or more features of the Calico wearable prototype and demonstrate how it can be used to augment one's life. These applications were all designed to fit into our generalized track layout.

### 6.1 End-User Auscultation

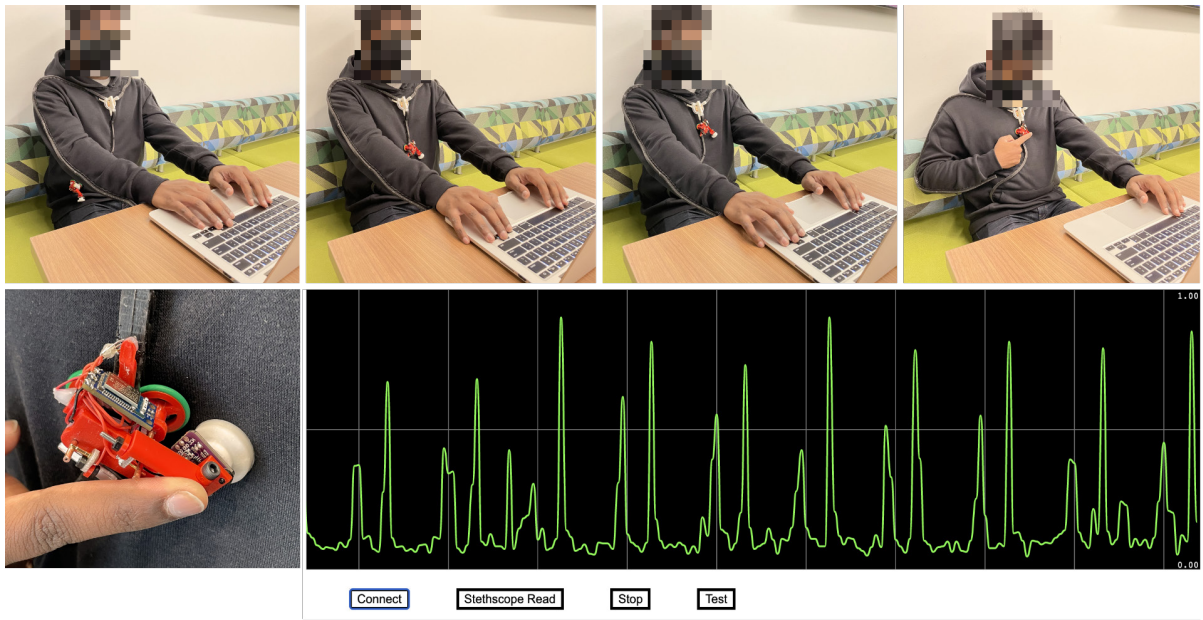

Fig. 18. Using Calico to perform a remote auscultation.

In this example, we showcase a scenario where a user can attach a sensor peripheral (an electret microphone with a stethoscope head) and collect on-demand chest auscultations in periodic intervals. Contact-free auscultations to regularly monitor patients has the potential to reduce the burden on the in-person healthcare system.

Although our technology is not mature enough to be deployed on any scale, we envision some benefits to providing a device that can move on the user's body when they are experiencing high fatigue or limited mobility. An upper body garment specifically designed to accommodate remote auscultations (Figure 18) can be worn by the patient. At regular intervals, an auscultation can be remotely administered by moving Calico to the right position and recording the audio signal. If necessary, a doctor can also control the movement of the robot. As the Calico system has magnetic landmarks, the doctor can monitor Calico's location with high precision. The user only has to apply light pressure to the top of the stethoscope head in order to record audio. The signal is then sent to the doctor in real-time (Figure 18). The scope can record raw audio at 10 ksps.

This application highlights the advantage of having a precise localization system and a high payload capacity that can be utilizied to attach bulky sensors and peripherals.

## 6.2 Workout Tracker + Form Coach

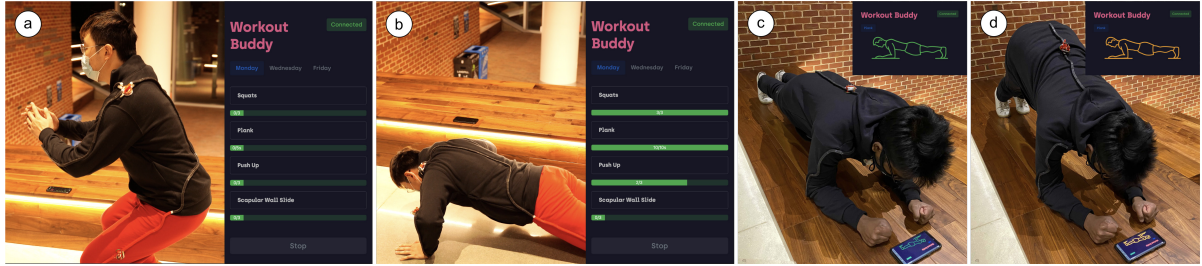

Fig. 19. Calico can function as a workout tracker and also a form coach. a) A user tracks their workout using Calico. When the user performs a squat, Calico moves to the knee to collect data and logs it in the app. b) Towards the end of their workout, the user performs a push-up, hence Calico moves to the elbow. c) Calico as a form coach monitoring the plank. d) When the user makes a mistake, Calico gives immediate feedback to the user.

Here, we showcase one mobile app that, together with Calico, can help the user monitor various exercises (Figure 19). When starting a training session, the user can pre-select a workout routine, for example, three squats (Figure 19a), five second of planks and then three push ups (Figure 19b). During the workout, Calico monitors the progress of the exercise, and moves to the appropriate location (knee for squats, back for plank and elbow for push ups) automatically. All this information is then relays back to the phone app. Due to Calico's high speed and robustness, the transition between exercises can happen very quickly without the user's intervention.

Besides tracking the user's workout, the relocatable property of Calico further allows it to monitor the performance of an exercise with a high accuracy, which can in turn help the user fix their form. Thus, Calico provides an extra feedback loop that allows the user to build muscle memory. For example, while performing a plank, core activation is largely affected when the performer doesn't maintain a neutral spine [50]. But as exercise is often a solitary activity, it's hard to receive the necessary feedback to correct this behavior. In Figure 19c and d, when the user selects a plank, the Calico robot will move to the user's back — the most effective body location for coaching a plank. When the user starts performing the plank, Calico moves back and forth on the user's back. The IMU readings are used to determine if the user's spine is neutral, which is essential for the plank. If the back sags or over-extends, the robot will notify the user with a vibration — generated by quickly moving the robot back and forth at micro-steps. The phone application also has the option to simultaneously remind the user (Figure 19d). As different users might have different form issues for different exercises, Calico can relocate to all the major points required to ensure that the exercise is performed with the right form.

This highlights Calico's use as a single wearable to collect various metrics. As various exercises are performed, Calico foregoes the necessity of attaching multiple sensors to multiple locations.

## 6.3 Dance-Dance Companion

In this example, we physicalize a fun application for Calico that cannot be supported by previous relocatable wearable systems. As Calico presents us with a wearable that can relocate on the body very quickly and precisely, it allows it to function as a dance trainer using IMU data.

For example, if a user wants to learn how to perform an "arm-wave", Calico can move along the arm while maintaining timing and rhythm as it can localize itself accurately while also moving without restricting the user's movement. If the goal arm wave takes 3 seconds from the shoulder to the wrist, Calico moves along the arm while simultaneously collecting IMU data along with tracking it's current location. If the user mistimes the movement, the IMU data indicates that the user has made a mistake (the gyroscope records the wrong rotation) and Calico can recognize that a mistake has been made and return to the starting location. This allows Calico to train the user to perform an arm-wave with the right timing. If the user is already a well-versed dancer, Calico can also serve as a dance companion as Calico can move in-sync with the dancer. A furry cap with googly eyes can be further installed to Calico, to make the dance companion a cute and unique creature (Figure 21).

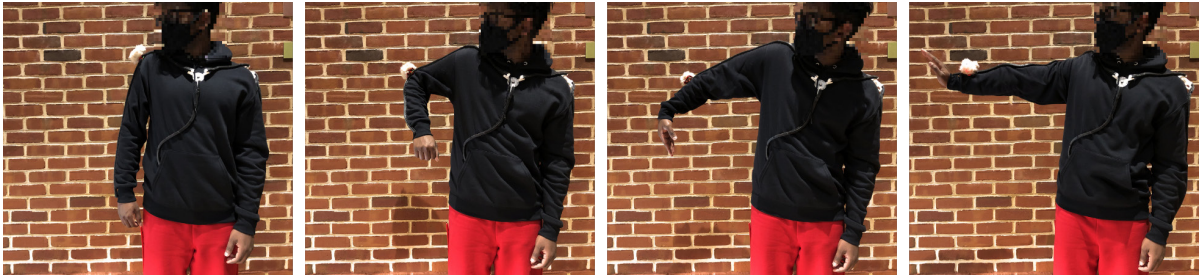

Fig. 20. Performing a dance with the on-body companion. The high speed and climbing reliability of Calico allows us to explore such applications.

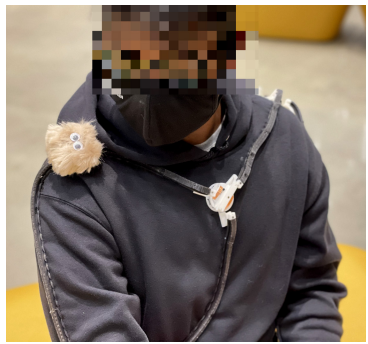

Fig. 21. A furry cap is used to make Calico more personable.

This highlights Calico's major contribution which is its ability to move without any restrictions to the user's movements and additionally using the user's movement to serve a purpose.

### 6.4 Data Physicalization Assistant

A function that is often provided by a digital assistant is the ability to store and display personal data that is required by the user. Let's say that a user is tracking the amount of water that they need to drink throughout the day. Usually, a phone app is used to log the water and the user also checks how much water they have already drank that day.

If we translate this functionality to Calico, we get certain additional benefits. Every time the user drinks water, they can "tap" Calico in a certain pattern to log this data. Additionally the user can also assign a segment of track to display this data. As shown in Figure 22, if the arm is assigned to track water intake data, this graph is physically projected onto the arm *i.e.*, Calico stays on the wrist to indicate 0 glasses and moves to the shoulder to indicate 6 glasses of water. During the day, as the user drinks more and more water, Calico moves up towards the shoulder. At any point of time, the user can immediately understand the amount of water that has been drunk by looking at the position of Calico instead of relying on an additional display to do so.

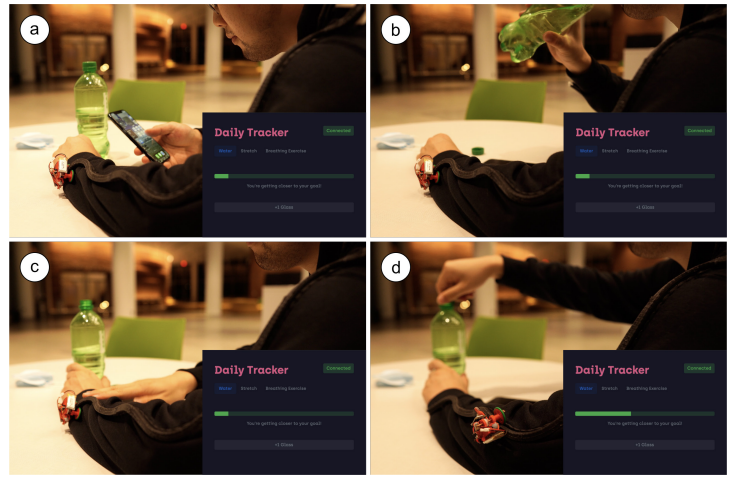

Fig. 22. Calico provides a tangible display for data. a) A user would like to track their water intake. b) They drink the water. c) They indicate to Calico that they have done so. d) Calico moves up along the designated track segment to physically "display" the data. The change in the data is also reflected on the phone app.

Similarly, any kind of 1-dimensional data can be attached to Calico with different segments of track to provide a tangible method to record and display data to the user. Calico provides a body-centric way [7] for the user to interact with their personal data.

## 7 USER STUDIES

To further investigate Calico's potential and its use cases, we conducted two sets of user studies. Compared to static wearables, the dynamic nature of Calico and its ability to provide continuous feedback is a major departure. In order to test the potential of continuous feedback, the first set of studies consisted of lab experiments with novice dancers (N = 14) who learned to perform an arm wave following the guidance of Calico vs. static vibration motors. In order to perform future research in this area, it is important to understand users' perceptions of such a system as they are not currently commonplace. The second set of studies consists of a survey to detect users' (N = 50) perceived usefulness, joyfulness, and social comfort regarding Calico in our application scenarios. The rest of this section details the procedure and the results of the user studies.

### 7.1 Lab Experiment: Using Calico to Assist the Learning of an Arm Wave

*7.1.1 Overview.* We designed a single-factor within-subjects experiment to investigate whether and how Calico could assist a novice dancer's learning of an arm wave (*i.e.*, the scenario detailed in section 6.3). Experiment sessions were conducted in the research group's lab space and were videotaped with the participant's consent. All participants performed three trials of the experiment task. During each trial, participants learned the arm wave with assistance until they felt ready, then performed the same arm wave independently. We manipulated the format of learning assistance across trials. Each participant was randomly assigned to one of the two tasks flows for counterbalancing:

- Video demonstration of the correct arm wave by an expert dancer. All participants received this format of learning assistance during their *first* trial, which mimics today's standard practice of how dance learning happens.

- Video demonstration, plus discrete vibration motors indicating the correct movement of the participant's arm. Half of the participants received this format of learning assistance during their *second* trial, while others received it in their *third* trial. The vibration motors measured a learner's arm position in real-time and compared it with the correct position that was supposed to happen at a given time point. Participants received on-body feedback as vibrational buzzes when their joints reach the correct positions (Figure 23).
- Video demonstration, plus Calico indicating the correct movement of the participant's arm. Half of the participants received this format of learning assistance during their *second* trial, while others received it in their *third* trial. Calico measured a learner's arm position in real-time and compared it with the correct position as displayed in the video. When a participant's arm hit the correct position that was supposed to happen at a given time point, Calico would start moving toward the next body part where the following position change should occur.

The above experiment design enabled us to detect the effect of Calico by comparing it with other learning assistance techniques for novice dancers (*i.e.*, the vibration motors). The adoption of a within-subjects design with video demonstration in all three trials helped minimize the confounding effect of individual differences in motor coordination.

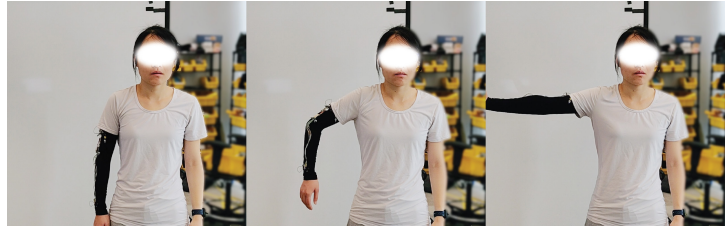

Fig. 23. A participant is using the vibration motors to learn an arm wave move.

*7.1.2 Participants.* We recruited 14 participants (6 females, 8 males) from a university in the United States. Their mean age was 26.64 years (SD = 9.11). Participants all identified themselves as novices at dancing. They self-reported moderate to high levels of ability to understand and use technology (M = 4.14, SD = 0.66 on a 5-point scale; 1 = very poor, 5 = excellent). Participants all performed three task trials but were randomly assigned to the two different task flows as described above.

*7.1.3 Measures.* We collected participants' self-evaluations and objective measures of the following aspects:

*Perceived Level of Difficulty of the Learning Task.* Participants reported their perceived level of difficulty of the arm wave learning task on a 5-point scale ("I believe I can perform this arm wave precisely without additional assistance."; 1= strongly disagree, 5 = strongly agree). This measure was collected upon a person's completion of the first task trial involving video demonstration only. Higher ratings indicated lower levels of task difficulty perceived by participants.

*Perceived Usefulness of the Learning Assistance Device.* Participants reported their perceived usefulness of the device (*i.e.*, vibration motors or Calico) on a 5-point scale ("The feedback from [the vibration motors or Calico] helped me learn how to perform this arm wave effectively."; 1= strongly disagree, 5 = strongly agree). This measure was collected after a person's completion of their learning processes at the second and the third task trials involving two different learning assistance devices, respectively. Higher ratings indicated greater levels of perceived device usefulness.

*Perceived Ease of Use of the Learning Assistance Device.* Participants reported their perceived ease of use of the device (*i.e.*, vibration motors or Calico) on a 5-point scale ("It was easy for me to understand the feedback provided by [the vibration motors or Calico].''; 1= strongly disagree, 5 = strongly agree). This measure was collected after a person's completion of their learning processes at the second and the third task trials involving two different learning assistance devices, respectively. Higher ratings indicated greater levels of perceived ease of device use.

*Perceived Joyfulness of Using the Learning Assistance Device.* Participants reported their perceived joyfulness of using the device (*i.e.*, vibration motors or Calico) on a 5-point scale ("I enjoyed having [the vibration motors or Calico] to help me learn how to perform this arm wave."; 1= strongly disagree, 5 = strongly agree). This measure was collected after a person's completion of their learning processes at the second and the third task trials involving two different learning assistance devices, respectively. Higher ratings indicated greater levels of perceived joyfulness of device use.

*Time Spent on Learning the Arm Wave.* The researcher used a stopwatch to calculate participants' time spent on learning the arm wave. At each task trial, a person's learning process started at the moment when they had been equipped with the learning assistance device and raised a hand as the sign of readiness to start learning. It ended at the moment when the person raised a hand again as the sign of readiness to stop learning and start performing. By the end of this calculation, each participant was matched to a learning time as in *the total number of seconds (s)*. Higher values indicated longer periods spent on participants' learning of the arm wave.

*Accuracy of the Arm Wave Speed.* The researcher performed a video analysis to calculate participants' speed of performing the arm wave and its accuracy against the correct speed as showed in the video demonstration. At each task trial, a person's performance started with the video frame where they issued the first arm movement. It ended with the video frame where the arm was fully stretched and held. By the end of this calculation, each participant was matched to a performance speed as in the total number of video frames (f). The accuracy of a person's performance speed was counted by *the distance in the number of video frames (f)* between their performance speed and the correct speed. Higher values indicated lower levels of speed accuracy.

*Accuracy of the Arm Positions.* For each task trial, the researcher calculated the moment-by-moment distance between the arm position angle performed by a participant and the correct angle as showed in the video demonstration. By the end of this calculation, each participant was matched to an accuracy score as in *the average number of distance in degrees (deg)*. Higher values indicated lower levels of position accuracy.

In addition, we interviewed all the participants to gain a richer understanding of their experience of using Calico in the context of dance learning. These interviews happened at the end of the experiment sessions. Each interview lasted for about 20 minutes.

*7.1.4 Results.* We built One-way ANOVAs models with Repeated Measures to compare the effects of different learning assistance devices on a person's self-reported device use experience as well as objective learning performance. All the models included participants' *perceived level of difficulty of the learning task* as the control variable. A Huynh-Feldt Correction was applied when the assumption of Sphericity was violated. We report the main effect of the learning assistance device (*i.e.*, the vibration motors vs. Calico) on each dependent variable. We also report the least squares means and standard errors for the tested variables.

Our first group of results considered the main effect of the learning assistance device on the participants' device use experience. The analysis indicated no significant main effect for the *perceived usefulness* (F [1, 12] = .01, $p$ = .91), *ease of use* (F [1, 12] = .19, $p$ = .67), and *joyfulness* (F [1, 12] = .15, $p$ = .71) across devices. Figure 24 illustrated the means and standard errors for each aspect of the device use experience measures. Specifically, there was a consistent pattern where Calico emerged to outperform the vibration motors in terms of its *perceived usefulness* ($M_{vib}$ = 3.71, $S.E._{vib}$ = .30; $M_{Calico}$ = 4.21, $S.E._{Calico}$ = .23), *ease of use* ($M_{vib}$ = 4.21, $S.E._{vib}$ = .23; $M_{Calico}$ = 4.57, $S.E._{Calico}$ = .17), and *joyfulness* ($M_{vib}$ = 3.93, $S.E._{vib}$ = .29; $M_{Calico}$ = 4.07, $S.E._{Calico}$ = .23) in assisting dance learning. Nevertheless, none of the differences reached significance.

Further, we tested the main effect of the learning assistance device on the participants' learning performance. The analysis indicated no significant main effect for the *time spent on learning* the arm wave: F [1, 12] = .33, $p$ =

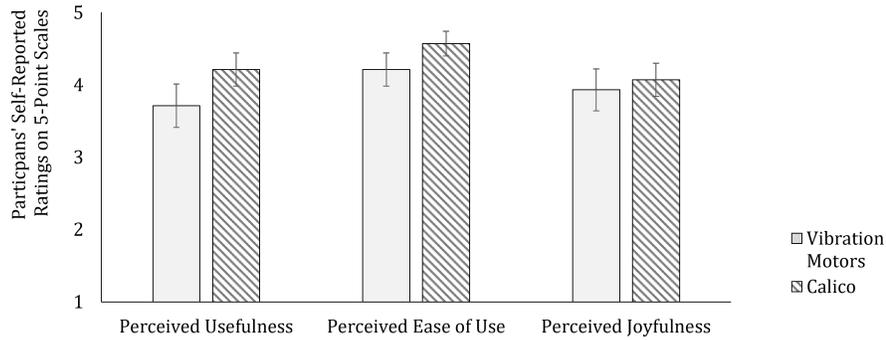

Fig. 24. Participants' self-reported ratings on the perceived usefulness, ease of use, and joyfulness of the two learning assistance devices.

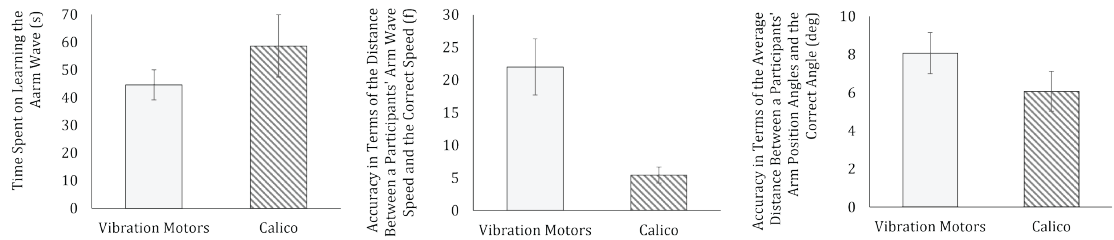

Fig. 25. The means and standard errors for each aspect of participants' learning performance after practicing the arm wave with devices.

.86. However, it suggested significant main effects for the *accuracy of the arm wave speed* (F [1, 12] = 13.46, *p* < .01) and the *accuracy of the arm positions* (F [1, 12] = 8.41, *p* = .01). Figure 25 illustrated the means and standard errors for each aspect of the participants' learning performance after practicing the arm wave with the devices. It took participants less *number of seconds to learn* the arm wave with the vibration motors ($M_{vib}$ = 44.57, $S.E._{vib}$ = 5.42) than with Calico ($M_{Calico}$ = 58.64, $S.E._{Calico}$ = 11.22) although the difference was not significant. Meanwhile, learning the arm wave with Calico resulted in significantly better performance outcomes in terms of less *delay of speed in the number of seconds (in frames)* ($M_{vib}$ = 22, $S.E._{vib}$ = 4.29; $M_{Calico}$ = 5.43, $S.E._{Calico}$ = 1.24) and less *distortion of arm positions in the average number of degrees* ($M_{vib}$ = 8.07, $S.E._{vib}$ = 1.08; $M_{Calico}$ = 6.07, $S.E._{Calico}$ = 1.05) against the standard.

Participants' interview responses highlighted a few unique strengths of Calico that were not specified in their ratings of the general user experience. In particular, 9 out of 14 participants pointed out that the *continuous feedback* provided by Calico was essential for dance learning. As explained by P2 and P6:

*"Calico definitely helped because I was able to follow it moment-by-moment. It's a good way to teach a person how to dance." [P2]*

*"With this robot, it felt more like there was a "guide" who kept showing me what position my hand should move towards next. The video and the vibration device didn't interact with me in a continuous way." [P6]*

For half of the participants, what they appreciated most was Calico's ability to provide *haptic feedback with perceived high resolution*. P5 and P9, for example, commented that neither the traditional video instructions nor the vibration motors could offer feedback of a comparable quality:

> "Calico was more helpful than other device [in teaching me the arm wave] because the feedback had higher granularity ... You can see how exactly the arm should be like just by following the trajectory." [P5]

> "The actual feel of the vibration device differed quite a lot from the robot. I think the movement of the robot helped me be aware of how my arm should be positioned. It provided richer haptic feedback." [P9]

Moreover, about one-third of our participants found that the *visual information* generated by Calico was as equally important as its haptic feedback and, sometimes, elicited the user's feeling of joyfulness. As said by P4 and P10:

> "I could see Calico was moving. That was good because I knew how fast my arm should be moving by monitoring the robot's movement." [P10]

> "Learning this dance move [with Calico] felt like playing catch-up by adjusting my arm position, which was fun." [P4]

Despite its strengths, several participants mentioned that it "felt a little weird wearing a robot gliding over the body"[P5], or it could be "awkward to have the robot crawl on my arm"[P14]. Some of them suggested tweaking Calico's aesthetics to provide a "pet-look" to enhance the users' adaptation to the robot, which echoes our idea of the furry Calico design illustrated in Figure 21:

> "It may be helpful to give Calico a creative and animal-like appearance ... If there were ways to make the robot look like a mini cat or a dog on the shoulder and let it move along my body, that would be cool." [P8]

In sum, the results of the lab experiments showed that Calico could enhance how novices learn dance moves by providing them with continuous feedback through both haptic and visual means. Although slightly better, the learners' perceived experience of interacting with Calico did not significantly differ from the experience of using discrete vibration motors. However, the use of Calico was associated with significantly improved learning outcomes in terms of less delay in the speed of the performance and less distortion of the arm positions.

### 7.2 Scenario-Based Surveys: Envisioning the Use of Calico in the Future

*7.2.1 Overview.* As described in Section 6, we envision that Calico would benefit users in future scenarios including and beyond dance learning. We designed scenario-based survey questions to detect peoples' reactions to the possibility of using Calico to assist end-user auscultation (*i.e.*, the scenario detailed in section 6.1), exercise coach (*i.e.*, the scenario detailed in section 6.2), and dietary data physicalization (*i.e.*, the scenario detailed in section 6.4). All participants were exposed to the three scenarios online. In each scenario, participants received a written description and a video to understand the challenge encountered by the protagonist and how Calico may help alleviate the challenge (*e.g.*, Figure 26). The use of a video allows our study to be more controlled as all participants witnessed the same demo. They then answered a set of questions asking about the perceived promises and constraints of Calico in that scenario. Leveraging the above study design, we gain the opportunity to investigate general users' understanding of the potential of Calico, especially in situations where the actual deployment of such device is not yet typical or feasible.

*7.2.2 Participants.* We recruited 50 participants (29 females, 21 males) through Prolific. Their mean age was 35.62 years (SD = 10.01). They self-reported moderate to high levels of ability to understand and use technology (M = 4.20, SD = 0.76 on a 5-point scale; 1 = very poor, 5 = excellent). Their average time to complete all the survey questions was 491.74 seconds (SD = 332.12).

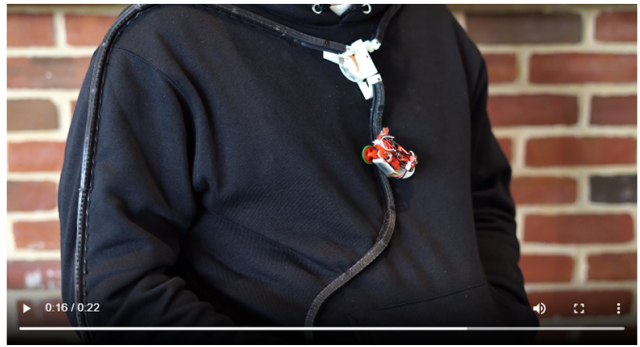

Fig. 26. An example of the scenario presented to survey respondents. In this example, participants received a written description and a video indicating the challenge of remote auscultation as well as how Calico may help alleviate the challenge.

*7.2.3 Measures.* We asked for participants' responses to an identical set of questions after reading each of the three scenarios. These questions measured the following aspects of a person's reaction to the possibility of using Calico in their future life:

*Perceived Usefulness*, *Ease of Use*, and *Joyfulness* of using Calico to address the challenge in each given scenario. We adopted the corresponding 5-point scales used in section 7.1.3 with modifications to fit the current study context. In the scenario of end-user auscultation, for example, participants indicated their level of agreement with the following statement (1 = strongly disagree, 5 = strongly agree): "If I were the patient in this scenario, I would be able to attend the auscultation with the help of this wearable robot," "If I were the patient in this scenario, it would be easy for me to understand how to interact with this wearable robot," and "If I were the patient in this scenario, I would enjoy having this wearable robot to help with the auscultation." Higher ratings indicated better qualities of user experience.

*Perceived Social Concerns* about using Calico with the presence of others in each given scenario. Wearable devices may elicit users' concerns, especially in non-private settings. Thus, we asked participants to identify the group(s) people in front of whom they would feel unconformable using the wearable robot. Participants indicated their choice(s) among six options, including family members, close friends, acquaintances, colleagues, strangers, and none (*i.e.*, always feeling comfortable using Calico in front of any others).

*7.2.4 Results.* We built One-way ANOVAs models with Repeated Measures to compare participants' perceived experiences of using Calico across three scenarios. A Huynh-Feldt Correction was applied when the assumption of Sphericity was violated. The analysis showed significant main effects of scenario on all aspects of user experiences, including the *perceived useful* (F [1.68, 98] = 22.56, $p < .01$), *ease of use* (F [2, 98] = 11.63, $p < .01$), and *joyfulness* (F [2, 98] = 30.91, $p < .01$). Figure 27 illustrated the means and standard errors for each aspect of the user experience measures.

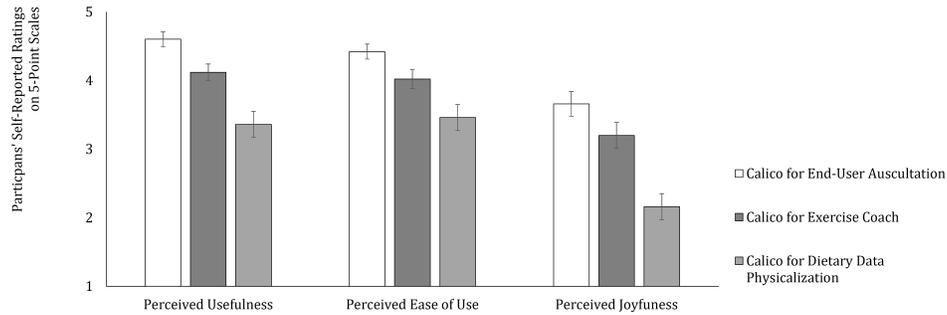

Fig. 27. The means and standard errors for each aspect of the user experience measures across scenarios.

Participants perceived Calico to be most useful in assisting end-user auscultation ($M_{ausc}$ = 4.60, $S.E._{ausc}$ = .11), which was followed by exercise coach ($M_{exer}$ = 4.12, $S.E._{exer}$ = .12) and dietary data physicalization ($M_{diet}$ = 3.36, $S.E._{diet}$ = .19). They found it would be easiest to use Calico for end-user auscultation ($M_{ausc}$ = 4.42, $S.E._{ausc}$ = .11), which was followed by exercise coach ($M_{exer}$ = 4.02, $S.E._{exer}$ = .14) and dietary data physicalization ($M_{diet}$ = 3.46, $S.E._{diet}$ = .19). They believed that it would be most enjoyable to use Calico in the scenario of end-user auscultation ($M_{ausc}$ = 3.66, $S.E._{ausc}$ = .18), which was followed by exercise coach ($M_{exer}$ = 3.20, $S.E._{exer}$ = .19) and dietary data physicalization ($M_{diet}$ = 2.16, $S.E._{diet}$ = .19). Pairwise comparisons indicated significant differences in the corresponding means between every two scenarios.

Further, participants' reports of social concerns showed that about half of these people would have no problem using Calico in front of others under any given scenarios. Among those who may feel uncomfortable using Calico in front of others, people would consider it more appropriate to use Calico with their strong ties (*i.e.*, family members, close friends) than with weak ties (*i.e.*, acquaintances, colleagues, strangers). Figure 28 illustrated the percentages of participants who reported being uncomfortable using Calico with the presence of various groups of others and across scenarios.

Together, the results of the scenario-based surveys suggested that general users could recognize the potential of using Calico to address various life challenges despite concerns. When it comes to high-stake scenarios (*e.g.*, end-user auscultation), a person's willingness to leverage Calico for task-oriented purposes would be much qualified by their perceived social appropriateness of using this wearable.

## 8 DISCUSSION AND FUTURE WORK

We now discuss limitations and additional considerations for future system design.

### 8.1 Anatomical Localization

Currently, Calico is capable of highly precise on-track localization. The track layout is translated to anatomical location based on the graph layout entered by us. With Calico's on-board sensing capability, we believe that it is possible to derive the anatomical location automatically without explicitly entering it into the graph structure. To understand the feasibility of the idea, we ran a proof-of-concept experiment with 5 independent auto calibration

Calico: Relocatable On-cloth Wearables with Fast, Reliable, and Precise Locomotion • 136:27

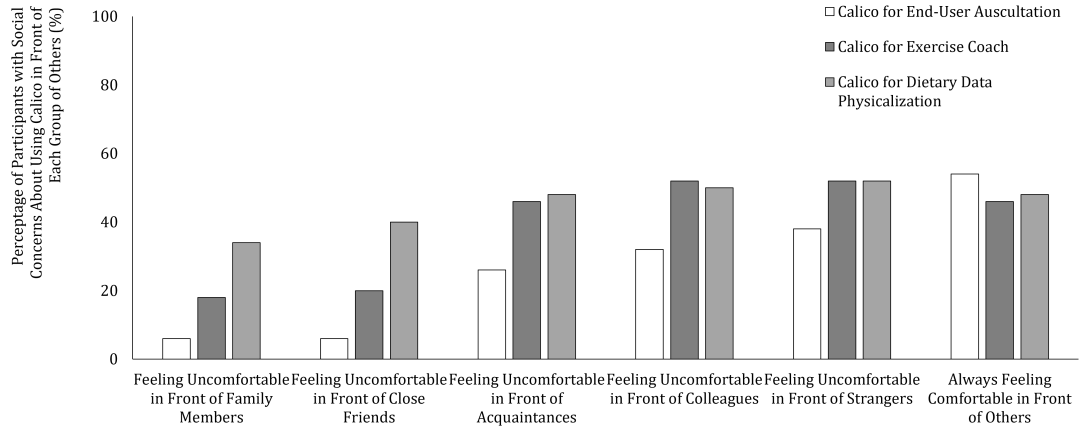

Fig. 28. The percentages of participants who reported being uncomfortable using Calico with the presence of various groups of others and across scenarios.

trials. For each trial, the user remains stationary in the standing posture with their hands holding the waist. The robot travels through the track from the user's wrist to the lower back while collecting sensor data from the IMU at 15 Hz and also counting the magnets that it crosses. The inertial sensor data collected in a fixed posture setting will be different at different anatomical location and thus the inertial sensor data can be used to dead reckon different anatomical locations. A single time or occasional auto-calibration step will allow the Calico robot to map magnet locations to the anatomical location and thus to dynamically adapt to changes.

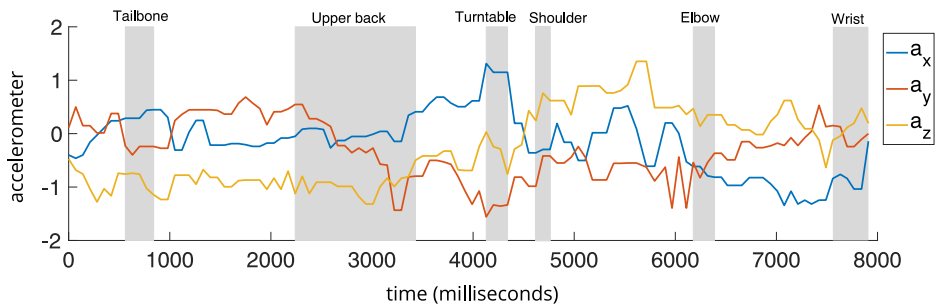

Fig. 29. The median filtered accelerometer time series data collected during a full body auto calibration step for precise anatomical localization.

Figure 29 illustrates the accelerometer data for one trial with high frequency noise removed. As seen in Figure 29, each anatomical location (with grey background) has unique signal trend. For instance, at the tailbone location, the robot senses positive linear acceleration along the x-axis and a negative value along y and z-axis; as the robot moves toward and then reaches the upper back area, the y-axis linear acceleration reaches its maximum and then drops sharply. The negative y-axis acceleration slope is a characteristic signature of upper back, which can be potentially used to identify this location. As the robot crosses the curvature of the shoulder, the robot registers a steady increase of the z-axis acceleration and a similar decrease in x-axis acceleration values. The crossing of these two signals (*i.e.*, x-axis and z-axis acceleration) happens right before the robot reaches the shoulder area. The turntable location gives rise to a sudden jump in accelerometer value along both x-axis and z-axis. These trends in inertial sensor data can be utilized to accurately recognize the anatomical locations.

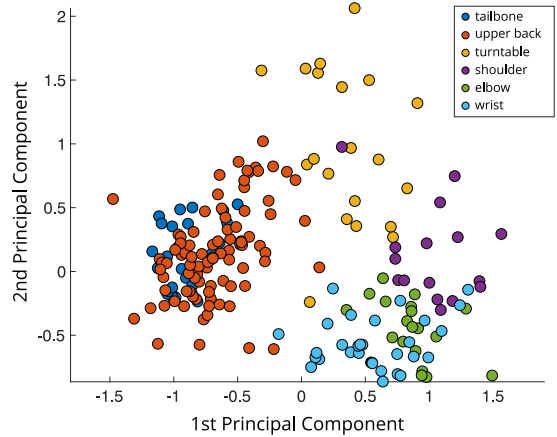

Fig. 30. shows that anatomical locations form clusters in the two dimensional space of the two principal components.

To further demonstrate the uniqueness of these trends for anatomical dead reckoning, we extract two features: the median filtered 3D accelerometer data and the median filtered data of the 3D accelerometer differentials. We then perform a Principal Component Analysis (PCA) on the 6 features corresponding to six locations (*i.e.*, tailbone, upper back, turntable, shoulder, elbow and wrist). Figure 30 shows a scatter plot of the projected accelerometer data. The data has formed clusters in different areas of the two dimensional space. For example, the data instances at the turntable and shoulder locations form clusters that are fairly different from the rest of the location clusters. While there are partial overlaps among the wrist and elbow clusters, the two clusters can be easily differentiated from the rest. Overall, the proof of concept experiment demonstrates the potential of a full-body auto calibration process to translate track location to anatomical location. For more rigorous evaluation, a more comprehensive data collection with a large and diverse cohort of participants is required.

## 8.2 Appearance and Styling

Since our system requires modifications to existing clothes with the add-on track system, it is important to examine the style and appearance of Calico. Here, we selected several common garment materials such as denim, faux leather, felt, satin, thick polyester with stripes of color, and a cloth with assorted patterns — this collection of fabrics presented us with a variety of textures, patterns, and shininess.

For each piece of cloth, we fabricated a 20 cm track unit. The tracks were dyed by mixing silicone pigments to match the appearance. Additionally, for the thick polyester with stripes of color, we created small sections of track pieces and glued them together with silicone epoxy. For the cotton with assorted patterns, we fabricated a translucent undyed track with an inlay made using translucent PETG filament. As shown in Figure 31, we can successfully match the track's color to most fabrics that come without strong reflections, although for satin, the track can be noticeable under strong lights. The clear track also has limited transparency and may partially block the patterns below.

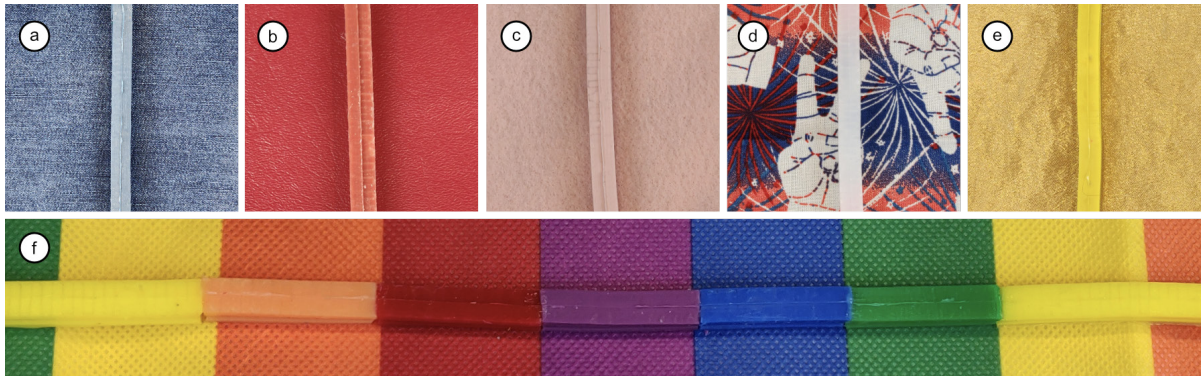

Fig. 31. a) Blue denim. b) Reddish Brown faux leather. c) Pink felt. d) Multi-colored pattern with a slightly translucent white track. e) Gold satin. As the fabric is extremely shiny, it is hard to blend with the fabric. f) Multiple colors of track epoxied together to form a single striped piece to match the pattern underneath.

### 8.3 Magnetic Landmarks

Calico is localized using embedded magnets inside the tracks that grant a minimum resolution of 4 mm. However, we should note that as the number of magnets increases, the track's flexibility and bending angle will be heavily affected. Alternatively, there might be some scope to replace both the magnets and the 3D printed live hinge with flexible magnetic tape. The alternating poles of the tape can also be detected by the Hall effect sensor. This will potentially reduce the fabrication complexity, as implanting magnets will no longer be needed.

### 8.4 The Future of Relocatable Wearables

Through the user studies, it is evident that using a relocatable wearable platform has interesting connotations. This is evident in the varied responses to how users perceive a relocatable wearable and the scenarios in which they deem it comfortable and socially acceptable. Some users are drawn to the joyful side of such a device, whereas others are drawn to the functional aspect.

This varied but interesting response allows us technologists to envision multiple variations and tangents. Leveraging the joyful side of the technology with animalistic "cute" behavior and aesthetics would allow us to explore how such technology can increase engagement with traditionally disengaging topics. On the other hand, the potential functionality of such a device with regards to accessibility or explorations like on-body memory aids and physical helpers could be impactful. Along with these, such a device also has the ability to relinquish control of an on-body device to a third-party agent. As resonated by P9 from the study in Section 7.1, this would allow us to explore thought-provoking questions regarding agency and control over our appearance and movement.

### 8.5 Ethical Considerations

As researchers, it is remarkably easy to focus on the novelty of the idea and the system implementation while disregarding the user's deep underlying motivations. Fit4Life [39] explores this idea by employing the use of a satirical system designed to help a user lose weight. Along the way, the system loses sight of the user's goals, and the data tracking mechanisms start to invade the privacy and the will of the user massively. As Calico aims to explore the concept of on-body wearables, it is important to consider the ethical implications of tracking personal data that align with the user's goals while maintaining user privacy. For example, tracking the posture of a user can be beneficial for their long-term health; using the same data to motivate the user's physical attractiveness is no longer aligned with the goals of the individual, and may affect the mindset of the user in the long term.

Extra care must be taken while implementing applications that involve personal companions. The effect of a digital companion on the user's long-term mental health must be thoroughly considered. Work like Jaron Lanier's [28] explores the evolution of personhood based on how computation is perceived. Extending the presence of computation to an always-on device on one's body with the capability to move around it can strongly influence the one's perception. Whether the influence has desirable connotations must be explored thoroughly before deploying such a system on any user.

## 9 CONCLUSION

We presented Calico, an interactive relocatable wearable using on-cloth track systems. We reported the design consideration and the implementation details of the Calico, with a series of technical evaluations to understand its performance. As Calico enabled fast, reliable, and precise on-cloth locomotion, we highlighted such capability with a suite of scenarios in which such a device would be useful. To understand how users would evaluate their interaction with Calico, we conducted lab experiments and scenario-based surveys. The lab experiments showed that Calico could help novices learn dance moves with its continuous feedback through haptic and visual means. The result of the scenario-based surveys suggested that general users could recognize the potential of using Calico to address various life challenges despite concerns.